\documentclass[10pt,twocolumn,letterpaper]{article}

\usepackage{cvpr}
\usepackage{times}
\usepackage{epsfig}
\usepackage{graphicx}
\usepackage{amsmath}
\usepackage{amssymb}

\usepackage[breaklinks=true,bookmarks=false]{hyperref}

\cvprfinalcopy 

\def\cvprPaperID{****} 

\begin{document}

\title{Visual Sensor Network Reconfiguration\\ with Deep Reinforcement Learning}

\author{
Paul Jasek \\
Ohio State University \\
Air Force Research Laboratory
\and
Bernard Abayowa \\
Air Force Research Laboratory
}

\maketitle

\begin{abstract}
We present an approach for reconfiguration of dynamic visual sensor networks with deep reinforcement learning (RL). Our RL agent uses a modified asynchronous advantage actor-critic framework and the recently proposed Relational Network module at the foundation of its network architecture. To address the issue of sample inefficiency in current approaches to model-free reinforcement learning, we train our system in an abstract simulation environment that represents inputs from a dynamic scene. Our system is validated using inputs from a real-world scenario and preexisting object detection and tracking algorithms. 

\end{abstract}

\section{Introduction}

The application of deep neural networks in reinforcement learning (RL) has shown success in a variety of domains. For example, Deep Q-Networks~\cite{Mnih2015} achieved human-level performance in Atari 2600 games. Other recent approaches, including trust region policy optimization~\cite{Schulman2015}, asynchronous advantage actor-critic~\cite{Mnih2016}, and proximal policy optimization~\cite{Schulman2017}, have shown success in domains such as 3D mazes and simulated robotic motion. However, the sample inefficiency of these algorithms limits the application of current deep RL solutions to many real-world problems where access to sample operations may be limited or expensive. A simulation environment generates sample observations quickly and cheaply, providing an RL agent with enough data to learn a high-performing policy.

We aim to apply reinforcement learning to a dynamic sensor-network configuration problem. While, we attempt maintain generality throughout our experiments, our specific motivation is to use cameras to capture high-resolution views of vehicles in a scene. Directly simulating this environment would involve a variety of difficult technical challenges and would likely be computationally expensive and unrealistic when compared to a real-world scenario. Instead, we focus on modeling an abstract scenario, where objects and sensors are represented as bounding boxes. A deep RL agent can learn to maximize the percentage of objects captured at high-resolution within a scene by training in this simulation environment. After learning an effective policy, the agent can operate within a real-world environment where preexisting object detection and tracking algorithms are applied to emulate the simulation environment from training.

\section{Related work}

Several methods have been proposed in the literature for reconfiguration of dynamic visual sensor networks of static and Pan-Tilt-Zoom (PTZ) cameras. These methods can be grouped into resource-aware methods, target-based methods, and coverage-oriented methods~\cite{piciarelli2016dynamic}.

Resource-aware methods seeks to find the optimal trade-off between available resources in the sensor network and task performance requirements. The sensing parameters are reconfigured to minimize usage of resources such as power in energy-aware surveillance systems~\cite{khan2012reinforcement} and communication bandwidth in distributed camera networks~\cite{karuppiah2010automatic}. Resource-aware methods are often found in settings where the visual sensors are static.

In target-based method the focus is on the optimization of the camera parameters to put a target of interest in view. Common applications include online adjustments of the orientation and zoom parameters of a PTZ camera for single target tracking~\cite{del2010exploiting}, and camera assignment or hand-off for optimal view in static camera networks. 

The group of methods related to ours are coverage-oriented methods in which the goal is to maintain optimal scene coverage with a network of PTZ cameras~\cite{kansal2007reconfiguration, konda2013optimal, piciarelli2011automatic}. In these methods, the parameters of the PTZ cameras are adjusted to maximize the view of relevant areas in the scene while also adapting to the scene dynamics. 

Existing methods for optimal coverage with visual sensor networks make use of hand-crafted mathematical models and shallow neural networks which do not generalize well. In this work we introduce a general framework for reconfiguring visual sensor networks to optimize coverage by leveraging advances in model-free RL and deep representation learning.

\section{Background}

This section contains relevant background information on the asynchronous advantage actor-critic (A3C) algorithm~\cite{Mnih2016} and the relational network (RN) module~\cite{Relational} which are the foundations of our solution. A3C is used as the reinforcement learning algorithm and training framework for our agent, while RN is used as part of the deep neural network architecture to enable the effective application of A3C to our specific problem.

\subsection{Asynchronous advantage actor-critic}

Consider the standard reinforcement learning scenario in which an agent interacts with an environment $\mathcal{E}$. At any given time step $t$, the agent receives a state $s_t$ and takes an action $a_t$ chosen from the set of possible actions $\mathcal{A}$ according to its policy $\pi$, which maps states $s_t$ to an action (or distribution of actions) $a_t$. The goal of the agent is to maximize the discounted return at any given state $s_t$, defined by $R_t = \sum_{k=0}^{\infty} \gamma^k r_{t+k}$, where $\gamma \in (0,1]$ is the reward discount factor. The value of any particular state $s$ following a policy $\pi$ is defined by $V^\pi(s) = \mathbb{E}[R_t|s_t = s]$. Similarly, the action value at any particular state is defined by $Q^\pi(s,a) = \mathbb{E}[R_t|s_t=s, a]$. These two quantities are used to define the advantage of an action $a_t$ in state $s_t$ given by $A(a_t,s_t) = Q(a_t, s_t) - V(s_t)$. The advantage function represents the expected increase in future reward if a given action is taken rather than following the current policy.

A3C is an example of a model-free policy-based method which trains an agent to maximize $R_t$ by updating the parameters $\theta$ of the policy $\pi(a|s;\theta)$. Methods stemming from the REINFORCE algorithm~\cite{Willia1992} update the parameters $\theta$ by performing approximate gradient ascent on $\mathbb{E}[R_t]$. The standard REINFORCE algorithm updates $\theta$ in the approximate direction of $\nabla_\theta \mathbb{E}[R_t]$ using the unbiased estimate $\nabla_\theta \log \pi(a_t|s_t;\theta)\mathbb{E}[R_t]$. Often, a function of the state known as the baseline $b_t(s_t)$ is subtracted from $R_t$ to reduce the variance of the estimate, while remaining unbiased. If $b_t(s_t)$ is learned estimate of $V^\pi(s_t)$, then $R_t - b_t$ can be seen as an estimate of the advantage $A(a_t, s_t)$, because $R_t$ estimates $Q^\pi(a_t, s_t)$ and $b_t$ estimates $V^\pi(s_t)$. 

A3C uses an estimate of the advantage to scale the policy gradient, $$A(s_t,a_t,\theta,\theta_v) = \sum_{i=0}^{k-1}\gamma^i r_{t+i} + \gamma^k V(s_{t+k}; \theta_v) - V(s_t;\theta_v).$$ Here, $V(s_t;\theta_v)$ is a learned estimate of the value function and $k$ varies across states, but is bounded above by $t_{max}$, the number of time steps performed before updating the policy parameters. To encourage exploration, an entropy regularization loss term, $H(\pi(s_t; \theta))$ is added to the objective function. Here, $H$ computes the entropy of a distribution. This adds an additional hyperparameter, $\beta$, which is used to scale the entropy regularization loss term. The resulting objective function for the policy is $\log \pi(a_t|s_t; \theta)A(s_t, a_t, \theta, \theta_v) + \beta H(\pi(s_t; \theta))$. To train the policy function $\pi$, we apply gradient ascent to this objective function with respect to the policy function parameters $\theta$. The estimated value function $V$ is trained via standard supervised learning to approximate the same bootstrapped estimate of $R_t$ that was used to compute the advantage function given by $\sum_{i=0}^{k-1}\gamma^i r_{t+i} + \gamma^k V(s_{t+k}; \theta_v)$.

The system is designed to be trained on multiple CPU cores by running parallel simulation environments for a fixed amount of time steps, $t_{max}$, before accumulating gradients and updating a global network. See the original paper~\cite{Mnih2016} for more details on how this is implemented.

\subsection{Relational Networks}

The RN module is designed with the capacity to reason about the pairwise relations in a set of objects. Consider a set of $n$ objects $O = \{o_1, o_2, ..., o_n \}$. Here, the $i^{\text{th}}$ object is an $m$-dimensional vector, $o_i \in \mathbb{R}^m$. Additionally, we consider a condition, $c \in \mathbb{R}^l$, represented as an $l$-dimensional vector. The RN is expressed as a composite function, 

\begin{equation}
\text{RN}(O,c) = f_\phi\left( \sum_{i,j} g_\theta (o_i, o_j, c) \right).
\label{eq:rn}
\end{equation}

Here $O$ and $c$ are defined as above and $f_\phi$ and $g_\theta$ are multilayer perceptrons (MLPs) with weights $\phi$ and $\theta$, respectively. In this formulation, the role of $g_\theta$ is to compute a relation vector corresponding to the relationship between two objects under a given condition. The role of $f_\phi$ is to construct an output based on all relations by operating on the sum of all relation vectors. 

Object representation vectors can be directly provided (as is the case for our inputs) or generated from another neural network module (such as a CNN) as demonstrated in~\cite{Relational}. The input size of $g_\theta$ is $2m + l$ and the network may be several layers deep. Naturally, the input size of $f_\phi$ must be equal to the output size of $g_\theta$ and may also consist of multiple layers. The result is a simple, end-to-end differentiable, neural network module that can effectively reason about object relations. 

\section{Deep Multi-view Controller}
\subsection{Problem formulation}

We consider a master-worker setup with a single stationary master camera which provides an overview of a scene of vehicles and multiple active cameras with a narrow field of view. Our goal is to view a maximum number of vehicles at a specified high-resolution. These vehicles may be moving or stationary and can exit or enter the scene at any point in time throughout the scenario. The scenario eventually ends, but the specific time that this happens is unknown to the agent.

We created an abstract simulation environment to enable the effective use of model-free reinforcement learning techniques. The developed simulation uses bounding boxes to represent the vehicles in a scene and the camera views. This abstraction generalizes the scenario to any sensor with a rectangular view and objects with similar movement patterns to vehicles. Objects within the simulation can randomly switch between moving and remaining stationary. Moving objects randomly turn by adjusting their direction continuously and may randomly reverse directions. The sensors within the simulation can select between five possible actions (do nothing, move up, move down, move left, and move right). The agent receives a positive reward whenever an object is captured at high-resolution by an active camera for the first time.

\begin{figure}[!htb]
\begin{center}
   \includegraphics[width=0.8\linewidth]{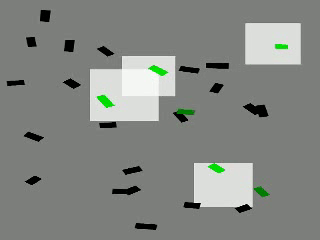}
\end{center}
   \caption{Simulation Environment used for training the agent. The light boxes represents sensor views within the environment. The green rectangles represent objects in the scene that have been captured at high-resolution, while the black rectangles represent objects that have yet to be captured at high-resolution.}
\label{fig:simulation}
\end{figure}

To increase the observability of environment, the simulation environment marks vehicles that have already been captured at high-resolution. This can be visualized in Figure~\ref{fig:simulation} where marked vehicles are represented as green. The simulation environment randomizes the number of objects and sensors, the object and sensor view sizes, the movement speeds of the sensors, and the time scale of observations by the agent. The purpose of this randomization is to increase the likelihood of a policy trained within the simulation environment being able generalize to a real-world scenario. We draw inspiration from recent work in which a robotic arm trained in a randomized non-photo-realistic simulation environment is able to perform the task in a real world setting without additional training~\cite{Tobin2017}.

The abstract representation was chosen, because we can use recent work in computer vision to translate a real-world scenario into the same representation. We assume access to sensor-view registration and an object and tracking system. While these systems are not trivial to implement, they are required to allow the agent to avoid keeping track of each previously detected object in the scene for the duration of the scenario. The absence of a tracking algorithm would require the agent to implicitly track each agent in the scene. Further, a system making use of the sensor network would likely be able to make use of an explicit tracking system for additional purposes.

\subsection{A3C with Multiple Agents}

Our agent is based on the A3C algorithm from ~\cite{Mnih2016}. We use multiple parallel simulation environments and update a global network by accumulating gradients computed from sample observations in each simulation environment. We enable the use of multiple sensors by controlling each sensor with a different instance of the same agent. We provide each instance of the agent with a similar input state, but mark a different sensor as the controlled sensor. In each simulation environment, we use the same agent to control each individual sensor, but only perform updates to the agent based on a chosen main agent in the simulation environment. The main agent receives credit for the reward received by all other agents within the environment. This was intended to increase cooperation between the agents in the scene by eliminating incentive for the agents to compete to capture the same vehicles at high-resolution.

\subsection{Model Architecture}

\begin{figure*}[!htb]
\begin{center}
\includegraphics[width=0.8\linewidth]{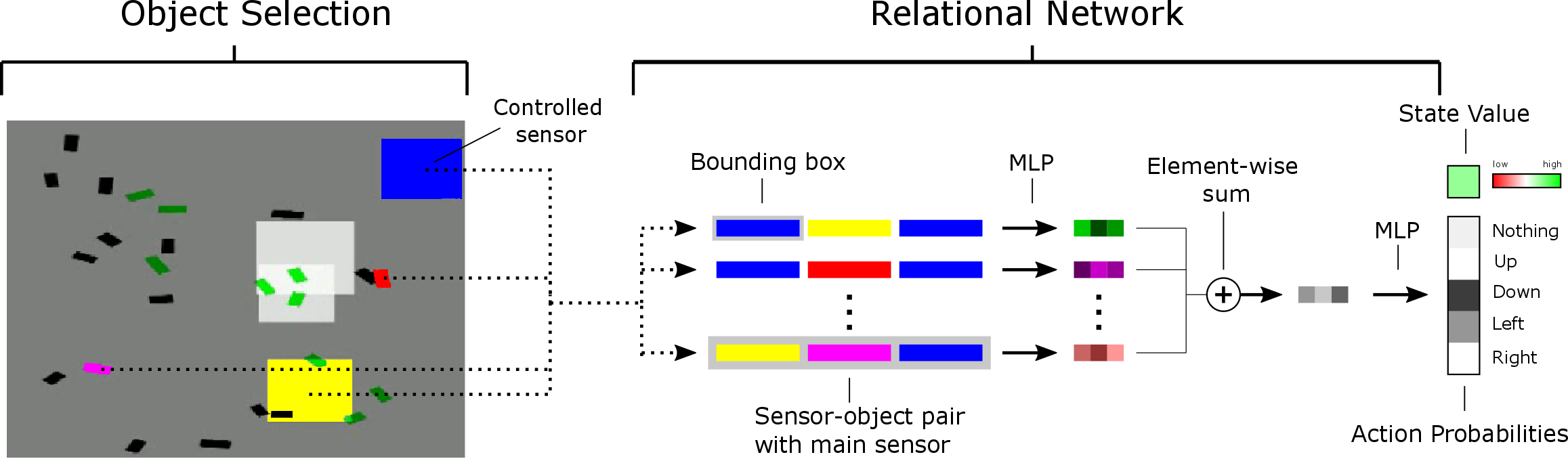}
\end{center}
   \caption{Graphical representation of network architecture used in our agent.}
\label{fig:architecture}
\end{figure*}

We base our network architecture on the A3C architecture used in~\cite{Mnih2016}. However, we make a significant modification by replacing the convolutional neural network (CNN) used to process the input state with a modified Relational Network (RN) module. 

The objects used by the RN are the object representations for each sensor and object in the scene from the last 4 time steps. Each object is represented by a 4-dimensional vector representing the bounding box of the object concatenated with a 1-dimensional vector representing type of object. Bounding boxes are represented as a vector containing the xy-coordinates of the object's center normalized to fall between -1 and 1 and the width and height of the object normalized to fall between 0 and 1. Concatenating these object vectors over the last 4 time steps results in 20-dimensional object representation vector. 

To optimize memory and computation time, we only consider relations between objects in which at least one of the objects is a sensor. Additionally, we do not show the agent objects which have already captured at high-resolution. The relations are conditioned upon the vector representation of the sensor that is being controlled by the agent. This results in a 60-dimensional vector representation for each relation. We pass each relationship vector through an MLP with 3 fully-connected layers with sizes 128, 256, and 256 respectively. This is the MLP represented by the function $g_\theta$ in Equation~\ref{eq:rn}. We then perform an element-wise sum operation and pass the result through a fully-connected layer with size 256 and 2\% dropout. Two separate fully-connected layers are applied to resulting output to produce a vector of five action probabilities and an estimate of value function. This is the MLP represented by the function $f_\phi$ in Equation~\ref{eq:rn}.

The entire architecture can be visualized in Figure~\ref{fig:architecture}. The element-wise sum operation in the RN module allows us to represent a dynamically sized input as a fixed-size vector. This vector is then used to generate the policy and value networks as done in the traditional A3C design. Arbitrary input size into the network is particularly important, because we have an arbitrary number of objects and sensors within the scene. In addition to feeling less natural, attempts to process our input using convolutional and recurrent neural networks resulted in significantly worse performance and increased running times.

\section{Training and Experiments}

\begin{figure}[!htb]
\begin{center}
\includegraphics[width=0.8\linewidth]{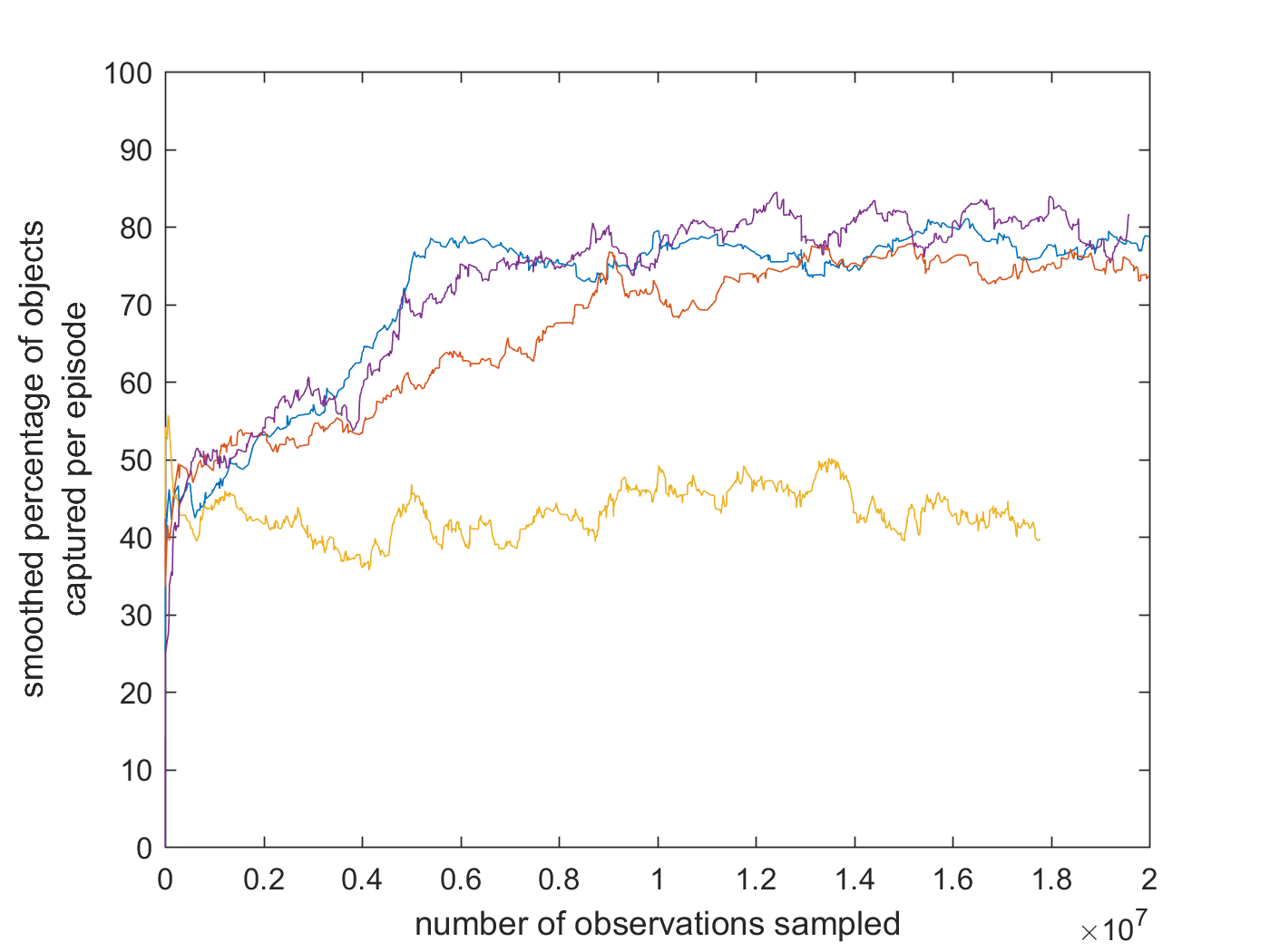}
\end{center}
   \caption{Smoothed performance over the course of training 4 agents.}
\label{fig:training}
\end{figure}

We trained out agent using 16 parallel simulation environments. The number of sensors and objects in each scene were selected from discrete uniform distributions with ranges of 1 to 5 and 1 to 50 respectively. We optimized three hyperparameters: learning rate, the entropy regularization constant, and the reward discount factor. We trained 4 agents with random hyperparameters from a limited range and selected the highest performing agent. A graph of this training process over the course of a week is shown in Figure~\ref{fig:training}. We observed instability in training with certain hyperparameters as can be seen with the yellow agent in Figure~\ref{fig:training}. The agent failed to learn a policy significantly better than simply taking random actions.

We evaluate our algorithm's performance by comparing with random movement and with a hand-crafted "lawn mower" method. We devised two random movement strategies which selected randomly and uniformly between the possible actions. One method included the "do nothing" action in which the camera simply remains in the same position for a time step, while the other method assigned zero probability to this action resulting in a slight performance increase. Under the "lawn mower" method, each active sensor view systematically covers the entire scene by moving up and down in columns, moving to the side for a single time step after reaching the top or bottom of the scene. The performance achieved by each method are shown in Table~\ref{tab:evaluation}. We can see that our agent performs significantly better than both the random and "lawn mower" strategies. Note that it certain scenarios within the simulation environment it may be impossible to capture 100\% of the objects at high-resolution, because objects may move out of the scene before any sensor is able to view it. We do not place a large emphasis on the specific percentage of vehicles captured, as the performance of the agent can be easily manipulated by adjusting the parameters of the environment such as the movement speed and view size of the active cameras. 

\begin{table}
\begin{center}
\begin{tabular}{|l|c|}
\hline
Agent & Percentage of Objects \\& Captured at High-Resolution \\
\hline\hline
Random with "do nothing" & 44.75\% \\
Random & 46.28\% \\
"Lawn mower" method & 64.44\% \\
Ours (stochastic) & 82.74\% \\
Ours (deterministic) & 84.75\% \\
\hline
\end{tabular}
\end{center}
\caption{Percentage of objects viewed at high-resolution over 100 episodes for our agents and several baseline methods. The stochastic version of our algorithm samples actions from the generated distribution. The deterministic version simply selects the action with the largest probability in the generated distribution, resulting in a slight increase in performance. }
\label{tab:evaluation}
\end{table}

\begin{figure}[!htb]
\begin{center}
\includegraphics[width=1.0\linewidth]{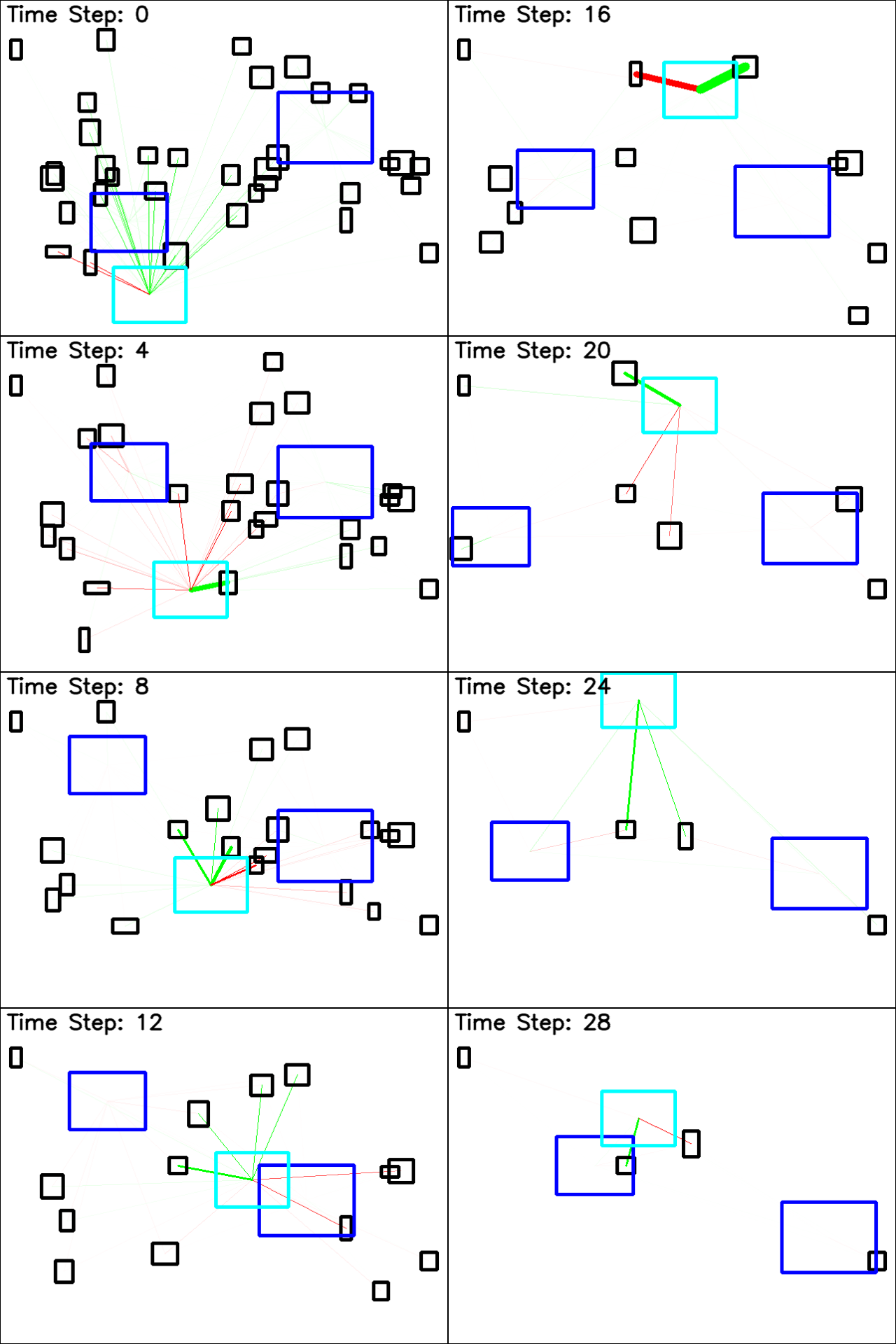}
\end{center}
	\caption{Visualization of the contribution of the relation between pairs of objects towards the chosen action under the learned policy. Relationships that contribute strongly to the chosen action are shown in green, while relationships that contribute negatively to the chosen action are shown in red. The controlled sensor is shown in cyan, while other sensors are shown in blue. The objects in the scene that have not yet been viewed at high-resolution are shown in black, while the objects that have are hidden.}
\label{fig:contributions}
\end{figure}

As an additional method of evaluating our learned policy, we look at the contribution of each relation considered by the agent towards the selected action distribution. This was calculated using the gradient of the KL divergence between the uniform action distribution and the selected action distribution with respect to a vector that is multiplied entry-wise with the computed relationships in the relational network evaluated at $\vec{1}$. Effectively, this results in a number scaled proportionally to the amount that each relation contributed to the chosen action distribution. The resulting contribution computation over several time steps in a scene can be visualized in Figure~\ref{fig:contributions}. Note that the sensor has learned to place high importance on the relations between itself and objects that are close to it. For example, at time step 16, we see that nearly all the focus of the policy is on the two objects closest to it. In this instance it chooses to captured the object on the left at high-resolution first. However, we see that the relationship between object to the right of the controlled camera contributes negatively to the agent's choice.

\begin{figure}[!htb]
\begin{center}
\includegraphics[width=1.0\linewidth]{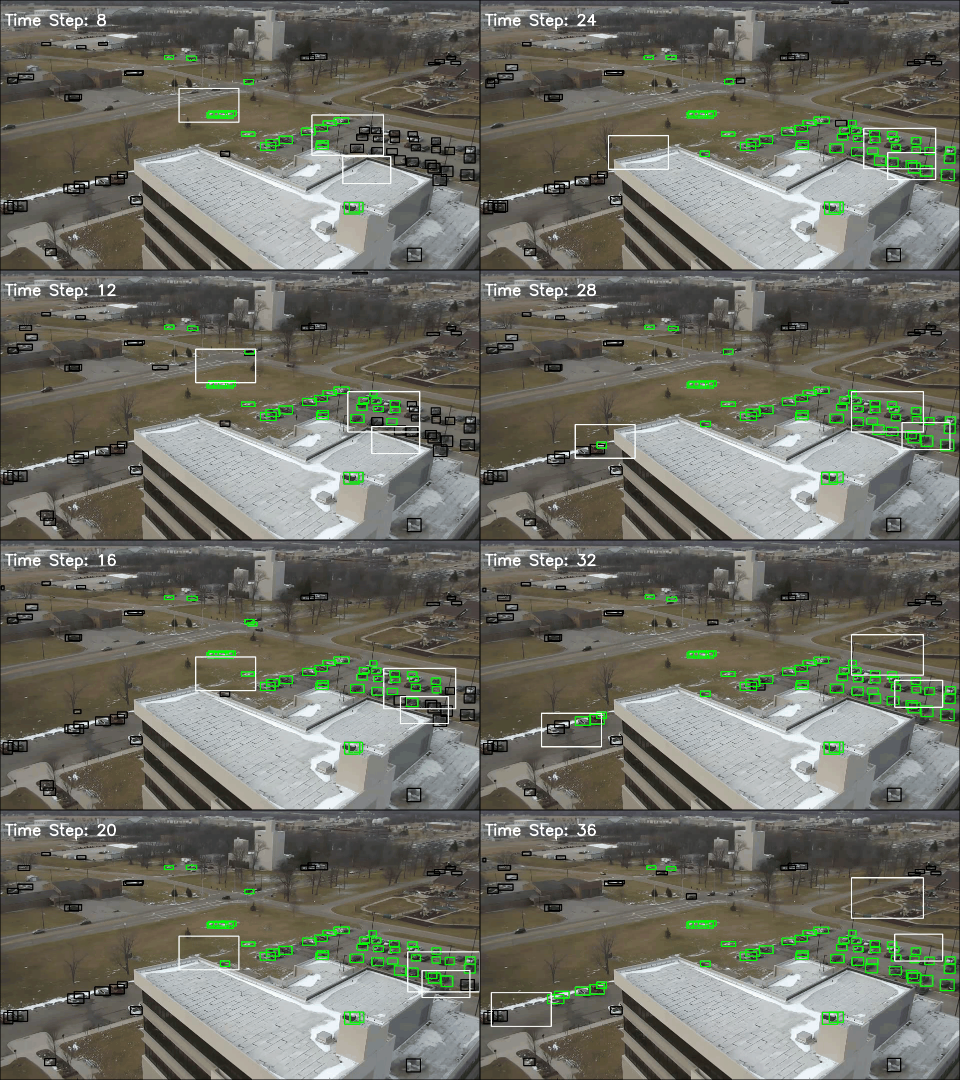}
\end{center}
	\caption{Visualization of the learned policy operating in a non-simulated environment. The white bounding boxes represent sensor views, while the vehicles in the scene that have not yet been viewed at high-resolution are shown in black, while the vehicles that have are shown in green.}
\label{fig:synthesis}
\end{figure}

We attempt to validate the ability of our learned policy to generalize by constructing pseudo real-world test environment. This involved a real-world video stream which was treated as the sensor reading from an overview camera. We applied real-time object detection and tracking to the video stream to simulate the inaccuracies in a real-world environment. The individual sensor views were simulated similarly to in the training simulation environment. Qualitative results to this experiment can be found in Figure~\ref{fig:synthesis}. Note that the object detector does not detect all vehicles and the tracker occasionally loses track of its targets.  This increases the difficulty of the reconfiguration task, while also providing a good test for the problems that may be encountered in a real-world environment.

\section{Conclusion}

We have shown that deep reinforcement learning can be applied to the problem of visual sensor network reconfiguration by training within a simulated environment. Although our results suggest that applying reinforcement learning to sensor network reconfiguration is feasible, there is much needed in terms of future work to reach a viable solution for the real world. There are several engineering challenges required to run object detection and tracking algorithms and control multiple cameras with low latency.

Additional future work is needed to improve the collaboration between multiple sensor controllers. Traditional reinforcement learning algorithms tend to ignore the case in which multiple agents are collaborating on a task. Our policy did not appear to learn many collaborative strategies outside of typically not overlapping sensor views. We can verify this by observing that the relation between the controlled sensor and other sensors in the scene does not seem to contribute to the selected action distribution as shown in Figure~\ref{fig:contributions}. Applying methods similar to recent research in multi-agent reinforcement learning such as MADDPG~\cite{Lowe2017} may increase the performance of the sensor network by increasing the level of collaboration between individual sensors in capturing all vehicles at high-resolution.

{\small
\bibliographystyle{ieee}
\bibliography{egpaper_final.bib}
}

\end{document}